\documentclass[10pt,twocolumn,letterpaper]{article}

\usepackage[pagenumbers]{cvpr} 
\usepackage{pifont} 
\usepackage[accsupp]{axessibility} 
\definecolor{cvprblue}{rgb}{0.21,0.49,0.74}
\usepackage[pagebackref,breaklinks,colorlinks,allcolors=cvprblue]{hyperref}

\newcommand{\cmark}{\ding{51}} 
\newcommand{\xmark}{\ding{55}} 

\def\paperID{37754} 
\def\confName{CVPR}
\def\confYear{2026}

\title{Unposed-to-3D: Learning Simulation-Ready Vehicles from Real-World Images}

\author{
Hongyuan Liu$^{1}$ \quad 
Bochao Zou$^{1}$\thanks{Corresponding author} \quad 
Qiankun Liu$^{1}$ \quad 
Haochen Yu$^{1}$ \quad 
Qi Mei$^{1}$ \quad 
Jianfei Jiang$^{1}$ \quad 
Chen Liu$^{2}$ \\ 
Cheng Bi$^{2}$ \quad 
Zhao Wang$^{2}$ \quad 
Xueyang Zhang$^{2}$ \quad 
Yifei Zhan$^{2}$ \quad 
Jiansheng Chen$^{1}$ \quad 
Huimin Ma$^{1}$\footnotemark[1] \\
\noalign{\vspace{5pt}}
$^{1}$University of Science and Technology Beijing \qquad $^{2}$Li Auto Inc.
}

\begin{document}

\twocolumn[{%
\renewcommand\twocolumn[1][]{#1}%
\maketitle
\vspace{-0.4cm}
\begin{center}
    \centering
    \includegraphics[width=\linewidth]{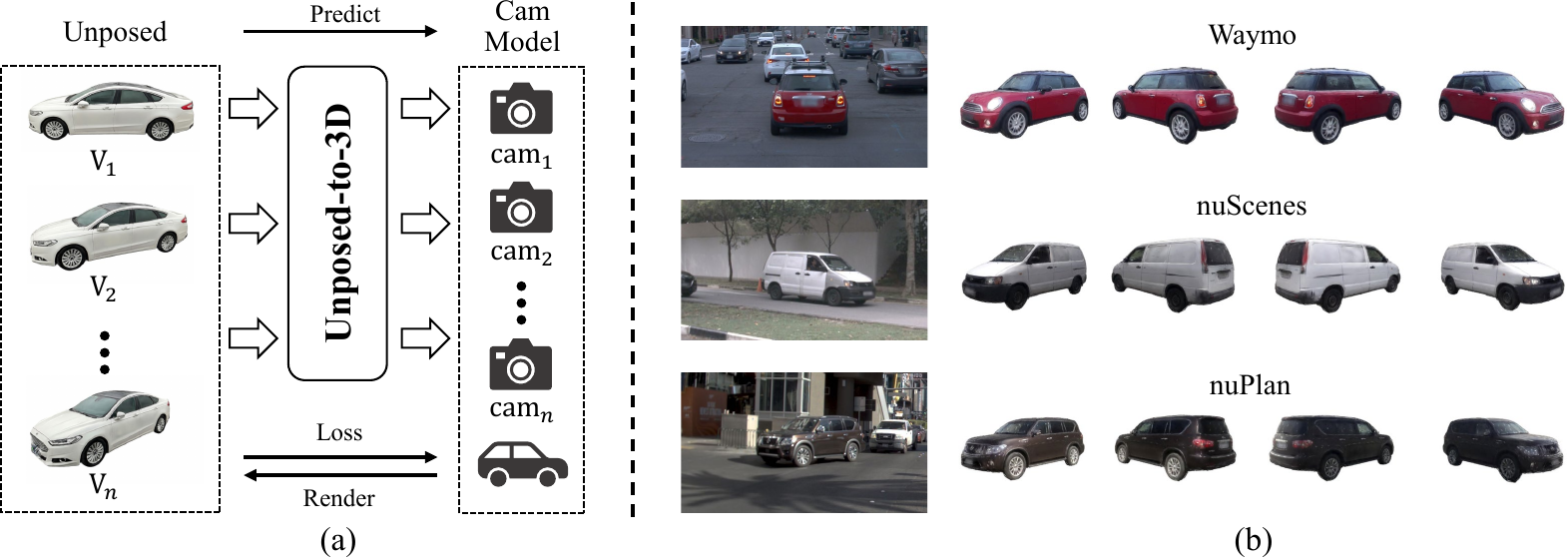}
    \vspace{-0.4cm}
    \captionsetup{type=figure}
    \captionof{figure}{(a) Our method reconstructs 3D assets directly from real-world images without camera pose annotations. It jointly estimates camera parameters and 3D structure, rendering images under predicted poses to compute reconstruction losses. (b) Trained entirely on real-world data, the model achieves accurate geometry, robust camera estimation, strong cross dataset~\cite{waymo,nuscenes,nuplan} generalization, and produces metric scale, simulation-ready 3D assets.}
    \label{fig:fig1}
\end{center}%
}]
\let\thefootnote\relax\footnotetext{*Corresponding authors: \{zoubochao,mhmpub\}@ustb.edu.cn}

\begin{abstract}
Creating realistic and simulation-ready 3D assets is crucial for autonomous driving research and virtual environment construction. However, existing 3D vehicle generation methods are often trained on synthetic data with significant domain gaps from real-world distributions. The generated models often exhibit arbitrary poses and undefined scales, resulting in poor visual consistency when integrated into driving scenes. In this paper, we present Unposed-to-3D, a novel framework that learns to reconstruct 3D vehicles from real-world driving images using image-only supervision. Our approach consists of two stages. In the first stage, we train an image-to-3D reconstruction network using posed images with known camera parameters. In the second stage, we remove camera supervision and use a camera prediction head that directly estimates the camera parameters from unposed images. The predicted pose is then used for differentiable rendering to provide self-supervised photometric feedback, enabling the model to learn 3D geometry purely from unposed images. To ensure simulation readiness, we further introduce a scale-aware module to predict real-world size information, and a harmonization module that adapts the generated vehicles to the target driving scene with consistent lighting and appearance. Extensive experiments demonstrate that Unposed-to-3D effectively reconstructs realistic, pose-consistent, and harmonized 3D vehicle models from real-world images, providing a scalable path toward creating high-quality assets for driving scene simulation and digital twin environments.
\end{abstract}    
\section{Introduction}
Reconstructing foreground objects plays a pivotal role in building realistic and controllable driving simulations. With the rapid advancement of 3D reconstruction techniques ~\cite{nerf,3dgs,jiangmvsmamba}, the fidelity and efficiency of scene reconstruction have been significantly improved, greatly facilitating virtual driving scene reproduction and closed-loop simulation~\cite{mars,unisim,street,hugsim,yu2025xyzcylinder,liu2025instdrive}. However, despite these developments, foreground object reconstruction remains a critical bottleneck. When the camera viewpoint deviates from the training range, especially under large view changes, reconstructed foreground vehicles often suffer from severe rendering artifacts or geometry collapse, limiting their applicability in dynamic driving scenarios.

Although recent progress in object-level 3D generation~\cite{shapee,get3d,get3dgs,dreamfusion} and reconstruction~\cite{lrm,tgs,lgm,trellis,wen2025dycrowd} has achieved impressive results, these methods are still difficult to apply directly to real-world driving simulations. A fundamental limitation arises from the lack of accessible, high-quality real-world 3D assets. Consequently, most existing approaches rely on synthetic datasets~\cite{objaverse,khanna2024habitat}, leading to evident domain gaps in geometry and texture between generated models and real vehicles. Furthermore, these methods typically generate objects with arbitrary pose and scale, lacking real-world spatial consistency. As a result, models must be manually aligned and rescaled before integration into real scenes, which severely restricts the practicality of current 3D generation approaches. In addition, when pre-generated assets are inserted into target scenes, discrepancies in lighting, shading, and weather conditions often produce strong visual disharmony~\cite{dipir,madrive,r3d2}, further degrading realism and simulation fidelity.

To address these limitations, we leverage the inherent sequential nature of driving data—where each scene naturally provides multiple temporal frames of the same object and design a framework that learns directly from real-world imagery as illustrated in \autoref{fig:fig1}. We propose a fully image-based 3D vehicle generation framework, termed Unposed-to-3D, which reconstructs simulation-ready vehicles without requiring explicit 3D supervision. Specifically, we first pretrain the model using a small set of images with known camera poses and introduce a dedicated camera token that predicts the camera parameters of the input view. This enables us to extend the training to large-scale unposed vehicle images, where the predicted pose serves as self-supervised geometry guidance during differentiable rendering. Moreover, we introduce a scale token to infer real-world object sizes and adjust the model’s 3D orientation by converting vehicle motion into equivalent camera pose transformations, producing pose-consistent and metrically grounded 3D reconstructions. Finally, we define the visual inconsistency caused by inserting reconstructed assets into various environments as a harmonization problem and propose a lightweight modulation module that refines lighting and appearance to achieve realistic adaptation across scenes. 

In summary, our main contributions are as follows:
\begin{itemize}
    \item Image-only 3D learning framework that reconstructs vehicles from real-world sequence images without any 3D geometry supervision, enabled by a camera prediction based self-supervised learning strategy.
    \item Scale-aware reconstruction pipeline that predicts metrically grounded and pose-fixed 3D models suitable for direct integration into simulation environments.
    \item Lightweight modulation module that adapts reconstructed assets to varying lighting and scene conditions, ensuring realistic and visually coherent insertion into diverse driving scenes.
\end{itemize}

\section{Related Works}

\begin{figure*}[t]
    \centering
    \includegraphics[width=\textwidth]{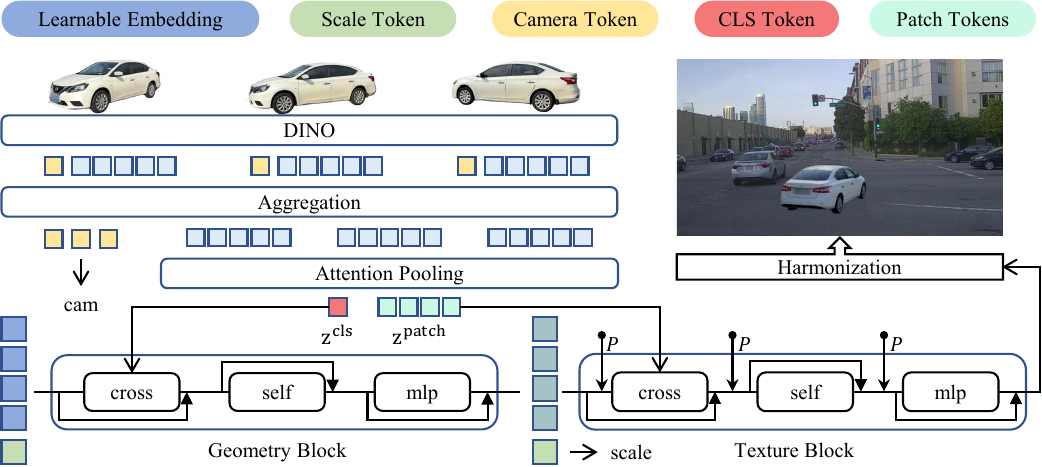}
    \caption{Given input images from arbitrary viewpoints, we first extract visual features using DINOv2~\cite{dinov2}. The Alternating Self-Attention module jointly infers camera parameters from a shared learnable camera embedding, while aggregating image features into a cls token and patch tokens via attention pooling. In the geometry block, the cls token is fused with learnable structure and scale embedding to reconstruct coarse geometry and predict scale. The texture block then refines appearance from patch tokens. The resulting metric scale 3D asset can be seamlessly inserted into novel scenes, where a harmonization module adjusts its appearance to match the surrounding context.}
    \label{fig:frame}
\end{figure*}

\subsection{3D Generative Models}
Early research on 3D generative modeling were largely constrained by the limited scale and quality of available 3D datasets~\cite{shapenet, pointe}. These methods are typically trained on small collections of meshes or voxel representations, resulting in low-fidelity geometry and texture quality that limited generalization to complex, real-world scenes~\cite{pointe,shapee,get3d}.
Leveraging 2D diffusion priors ~\cite{rombach2022high, saharia2022photorealistic}, DreamFusion ~\cite{dreamfusion} proposed Score Distillation Sampling (SDS) for 3D optimization. Follow-up studies ~\cite{liang2024luciddreamer, lin2023magic3d, dreamgaussian, wang2023prolificdreamer} have since enhanced both the efficiency and fidelity of this process.
Zero-1-to-3~\cite{liu2023zero} decomposes the 3D model generation task into multi-view image synthesis and sparse-view reconstruction. Subsequent works have extended this framework to different modalities, achieving impressive quality in 3D model generation~\cite{li2023instant3d,liu2024one,xu2024instantmesh}. However, such approaches that rely on image generative models remain constrained by the geometric consistency of the generated images.
More recently, large reconstruction and generative models~\cite{lrm,lgm,trellis} trained on massive, high-quality 3D datasets have demonstrated impressive generalization to unseen categories and high quality object synthesis. 

However, despite their success, such methods still rely heavily on the availability of curated, high-fidelity 3D datasets.
In contrast to these data-intensive paradigms, our work focuses on exploiting the abundance of 2D images to generate simulation-ready, high-quality foreground assets without the need of 3D supervision.

\subsection{Foreground Reconstruction}
Autonomous driving simulation has evolved from traditional CAD-based models to neural implicit representations such as NeRF~\cite{nsg,mars,unisim}, and more recently to 3D Gaussian Splatting~\cite{hugsim,street}, which enables high-fidelity real-time rendering. Several works~\cite{geosim, cadsim} attempt to reconstruct foreground vehicles from real-world driving scenes by leveraging point clouds and semantic segmentation maps. However, these approaches lack the ability to synthesize novel views with consistent appearance. Other methods~\cite{madrive, urbancad} retrieve 3D vehicle models from pre-built databases and replace detected objects with the best-matched templates. While effective in some cases, such non-learning-based pipelines are fundamentally limited by the diversity of their databases. NeRF-based approaches~\cite{autorf,carstudio,gina} generally suffer from low-quality textures and are computationally impractical for real-time simulation. ProtoCar~\cite{protocar} further introduces 3DGS for representing vehicle assets, and DreamCar~\cite{dreamcar} proposes a coarse-to-fine multi stage optimization framework. 

Nonetheless, these methods still depend on posed images and 3D supervision, while their visual fidelity remains limited. In contrast, by learning from purely image-level supervision, our method generates simulation-ready, high-quality 3D foreground assets, achieving both geometric accuracy and photorealistic texture.

\subsection{Pose Estimation}
Recent studies have explored recovering object poses directly from real-world images ~\cite{mousavian20173d}. Orient Anything ~\cite{wang2024orient} introduces a generalizable framework capable of predicting the 3D orientation of arbitrary objects. In autonomous driving, estimating the 6-DoF poses of vehicles is crucial for scene simulation and environmental perception ~\cite{xiang2015data,cfv,ke2020gsnet}. DUSt3R ~\cite{wang2024dust3r} achieves dense 3D scene estimation, while VGGT ~\cite{vggt} further extends this capability and explicitly predicts camera parameters, significantly advancing the development of pose-free scene reconstruction methods ~\cite{huang2025no}.

However, the intrinsic ambiguity between object geometry and camera parameters often leads to geometric collapse when the predicted camera parameters are directly used for 3D object generation. To overcome this challenge, we introduce a progressive optimization strategy: the model is first pre-trained on a small set of posed images, and subsequently fine-tuned on large-scale unposed datasets using image reconstruction losses. By jointly optimizing camera prediction and reconstruction consistency, our method achieves robust geometry estimation using only image-level supervision.

\begin{figure*}[ht]
    \centering
    \includegraphics[width=\textwidth]{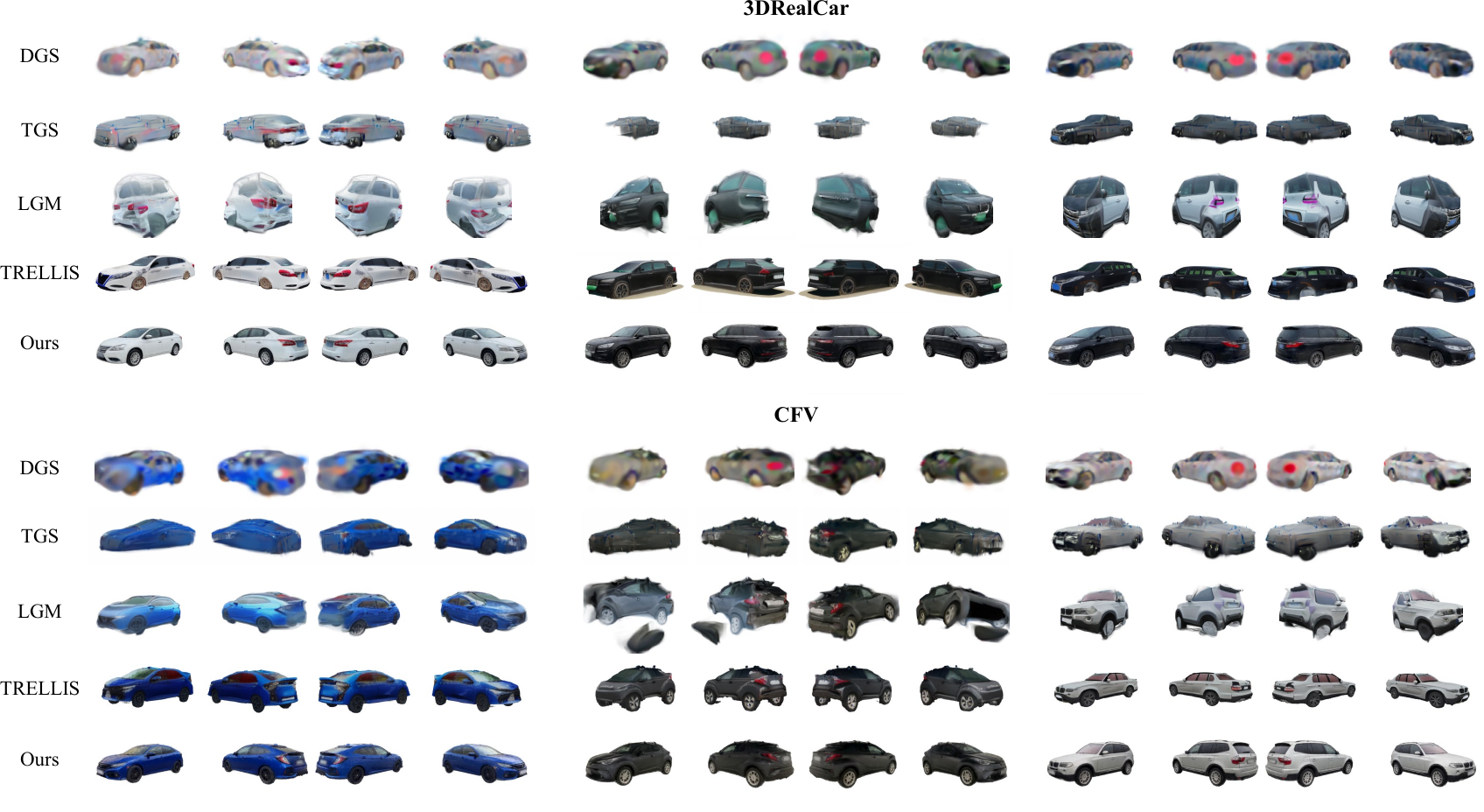}
    \vspace{-0.6cm}
    \caption{Qualitative comparison of reconstruction quality. For fair comparison, both our method and the baselines follow the single-image reconstruction paradigm. We present results on the 3DRealCar~\cite{3drealcar} and CFV~\cite{cfv} datasets. The baseline methods exhibit poor geometry and texture learning, and in some cases completely fail to produce valid reconstructions. In contrast, our model, trained purely on real-world image data, generates high-fidelity 3D assets across datasets, avoiding the synthetic and unrealistic artifacts introduced by simulation data.}
    \label{fig:show}
    \vspace{-0.2cm}
\end{figure*}

\section{Methodology}
Our framework is illustrated in \autoref{fig:frame}. We begin by employing the pretrained DINOv2 ~\cite{dinov2} to extract visual representations from the input image sequence. These features are subsequently fed into an aggregation module ~\cite{wang2024dust3r,vggt} designed to process sequential driving data, thereby enabling consistent 3D reconstruction from an arbitrary number of input views. To enable self-supervised learning on unposed images, we introduce a learnable camera embedding that predicts camera parameters for all input views, jointly optimized with the reconstruction process through image reconstruction losses in the second stage. The aggregated multi-frame features are then leveraged to guide geometry and texture learning: learnable structure and scale embeddings first recover coarse geometry and real-world scale, followed by decoding all Gaussian attributes. Finally, a lightweight modulation module adapts the representation to diverse scene environments, further enhancing the adaptability and generalization of our method.

\subsection{Visual Feature Extraction}

DINOv2~\cite{dinov2} encoder is employed to extract visual features from each input image $I_i$. Each image is then tokenized into a class token $\mathbf{x}^{\text{cls}}$ and patch features. Furthermore, we prepend a learnable camera token $\mathbf{x}^{\text{cam}} \in \mathbb{R}^{d}$ shared across all views to encode consistent pose-dependent features:
\begin{equation}
\mathbf{X}_i = \big[ \mathbf{x}^{\text{cam}}, \mathbf{x}_i^{\text{cls}}, \mathbf{x}_i^{1}, \ldots, \mathbf{x}_i^{K} \big] \in \mathbb{R}^{(K+2) \times d}.
\label{eq:dino_features}
\end{equation}

\begin{table*}[t]
\centering
\setlength{\tabcolsep}{4pt} 
\begin{tabular}{lccccc|ccccc}
\toprule
& \multicolumn{5}{c|}{\textbf{3DRealCar}} & \multicolumn{5}{c}{\textbf{CFV}} \\
\cmidrule(lr){2-6} \cmidrule(lr){7-11}
Method & SSIM↑ & PSNR↑ & LPIPS↓ & CD↓ & F-score↑ & SSIM↑ & PSNR↑ & LPIPS↓ & CD↓ & F-score↑ \\
\midrule
DGS      & 0.8635 & 19.6167 & 0.1442 & 1.3511 & 0.3022 & 0.8602 & 19.1443 & 0.1505 & 1.2446 & 0.3357 \\
TGS      & 0.9100 & 17.6330 & 0.1282 & 3.4874 & 0.2029 & 0.9048 & 17.1898 & 0.1320 & 3.3938 & 0.2157 \\
LGM      & 0.8412 & 17.8875 & 0.1450 & 7.1201 & 0.2030 & 0.8353 & 17.3817 & 0.1532 & 8.3786 & 0.1908 \\
TRELLIS  & 0.8879 & 19.5118 & 0.0884 &{1.3111} &{0.4744} & 0.8837 & 18.9785 & 0.0944 &{5.0039} &{0.3355} \\
Ours     & \textbf{0.9172} & \textbf{21.2033} & \textbf{0.0571} &\textbf{0.5782} &\textbf{0.5341} & \textbf{0.9183} & \textbf{21.7250} & \textbf{0.0508} &\textbf{0.7439} &\textbf{0.4252} \\
\bottomrule
\end{tabular}
\caption{Quantitative comparison on 3DRealCar and CFV datasets. For fair comparison, we follow a single-view inference paradigm and compute metrics across all views for both our method and the baselines. All baseline-generated models are manually aligned to canonical pose and normalized to standard scale, whereas our method requires no additional manual adjustment. Chamfer Distance (CD) is multiplied by $1\times10^3$, and F-score is computed with a threshold of $0.01$.}
\label{tab:single}
\vspace{-0.2cm}
\end{table*}

To integrate information both within and across views, an Alternating Self-Attention mechanism is adopted ~\cite{vggt}, where intra-frame self-attention enhances local texture details, and inter-frame self-attention captures holistic geometry, yielding camera features from $V$ views which are then decoded into camera parameters ${\theta}_i$ and aggregated features $\tilde{\mathbf{X}}_i$. For each image $I_i$, we predict the corresponding camera parameters:
\begin{equation}
{\theta}_i = { \mathbf{q}_i, \mathbf{t}_i, \mathbf{f}_i },
\label{eq:camera_params}
\end{equation}
where $\mathbf{q}_i \in \mathbb{R}^{4}$ denotes rotation, $\mathbf{t}_i \in \mathbb{R}^{3}$ is translation, and $\mathbf{f}_i \in \mathbb{R}^{2}$ represents the field of view. Following prior works~\cite{schonberger2016structure,wang2024vggsfm,vggt}, we assume that the camera has no principal point offset.
Considering the sign ambiguity of quaternions, the rotation loss is defined as $\mathcal{L}_{\text{rot}} = \min \left( | \mathbf{q}_i - \hat{\mathbf{q}}_i |_1, | \mathbf{q}_i + \hat{\mathbf{q}}_i |_1 \right)$ , and overall camera loss is given by:
\begin{equation}
\mathcal{L}_{\text{cam}} = \mathcal{L}_{\text{rot}} + | \mathbf{t}_i - \hat{\mathbf{t}}_i |_1 + | \mathbf{f}_i - \hat{\mathbf{f}}_i |_1.
\label{eq:camera_loss}
\end{equation}

The aggregated multi-view features $\tilde{\mathbf{X}}_i$ are then fused via attention pooling to produce global representations:
\begin{equation}
\mathbf{z}^{\text{cls}}, \mathbf{z}^{\text{patch}} = \sum_{i=1}^{V} \text{softmax}\Big(\text{MLP}(\tilde{\mathbf{X}}_i) \Big)\odot \tilde{\mathbf{X}}_i,
\end{equation}
where $\mathbf{z}^{\text{cls}}$ encodes coarse geometric structure and $\mathbf{z}^{\text{patch}}$ preserves fine texture information.

In the second stage, we supervise solely with the image reconstruction loss. Leveraging the differentiable rendering of 3DGS, the gradient of the image loss with respect to the camera parameters is backpropagated through each Gaussian's pixel position $\mu_i$ and covariance $\Sigma_i$:
\begin{equation}
\frac{\partial \mathcal{L}}{\partial \theta}
= \sum_i \Big(
\frac{\partial \mathcal{L}}{\partial \mu_i}^\top 
\frac{\partial \mu_i}{\partial \theta}
+ 
\big\langle 
\frac{\partial \mathcal{L}}{\partial \Sigma_i},\frac{\partial \Sigma_i}{\partial \theta} \big\rangle\Big), 
\end{equation}
where $\theta$ denotes the camera parameters. The camera extrinsics $(\mathbf{q},\mathbf{t})$ are primarily driven by the projection center errors, while the intrinsics $\mathbf{f}$ are influenced by both the mean and the covariance.

\subsection{Structure Feature Learning}


Directly mapping visual features to 3D representations is inherently challenging. We therefore decompose this process into two stages: coarse geometry prediction and fine-grained texture generation. While prior works often employ triplane or similar compression strategies~\cite{lrm,tgs,gina} to reduce the complexity of spatial feature construction, such approaches inevitably lead to information loss and prevent full exploitation of the learned features. 

To preserve detailed structure, we explicitly construct a set of $N$ learnable embedding tokens $\mathbf{E} = \{\mathbf{e}^1, \mathbf{e}^2, \ldots, \mathbf{e}^N\}$, each corresponding to a coarse geometric location, along with an additional scale token to regress the real-world scale. As illustrated in ~\autoref{fig:frame}, the Geometry Block consists of a sequence of cross-attention, self-attention, MLP, and with skip-connection. We incorporate geometric priors using a single learnable cls token, which captures rich global geometric information sufficient for decoding coarse geometry while substantially reducing computational cost.

The scale token is decoded by a scale prediction head into a spatial bounding box $b \in \mathbb{R}^3$, representing the object size along three axes. The geometry features $\mathbf{G}_i$ encoded by the Geometry Block are further decoded by a position prediction head to obtain the spatial position of each token $P_i = (x_i, y_i, z_i)$. In practice, due to the absence of explicit 3D supervision, directly predicting $(x, y, z)$ from image losses is extremely difficult. To address this, we propose a progressive optimization strategy that transforms the positional representation from Cartesian coordinates to spherical coordinates $(r, \theta, \phi)$. Given that the object center is constrained at the world origin, we first optimize the radial distance $r$ during early training, and subsequently optimize the azimuth angle $\theta$ and elevation angle $\phi$. This coordinate transformation substantially reduces the complexity of spatial learning without sacrificing representational accuracy. 

Tokens at different spatial positions should exhibit distinct texture characteristics. Therefore, we employ the predicted position $P_i$ to modulate the corresponding geometry feature $\mathbf{G}_i$ through adaptive layer normalization (adaLN)~\cite{dit}. Specifically, the modulation parameters are predicted from $P_i$ as $\gamma_i, \beta_i = f_{\mathbf{mod}}(P_i)$, which control the scaling and shifting of each feature dimension, respectively:
\begin{equation}
{\mathbf{T}}_i = (1+\gamma_i) \cdot \text{LN}(\mathbf{G}_i) + \beta_i,
\end{equation}
where $\text{LN}(\cdot)$ denotes layer normalization, and ${\mathbf{T}}_i$ represents the modulated texture feature that incorporates spatial awareness and facilitates consistent geometry-texture alignment in subsequent stages.

\begin{figure*}[ht]
    \centering
    \includegraphics[width=\textwidth]{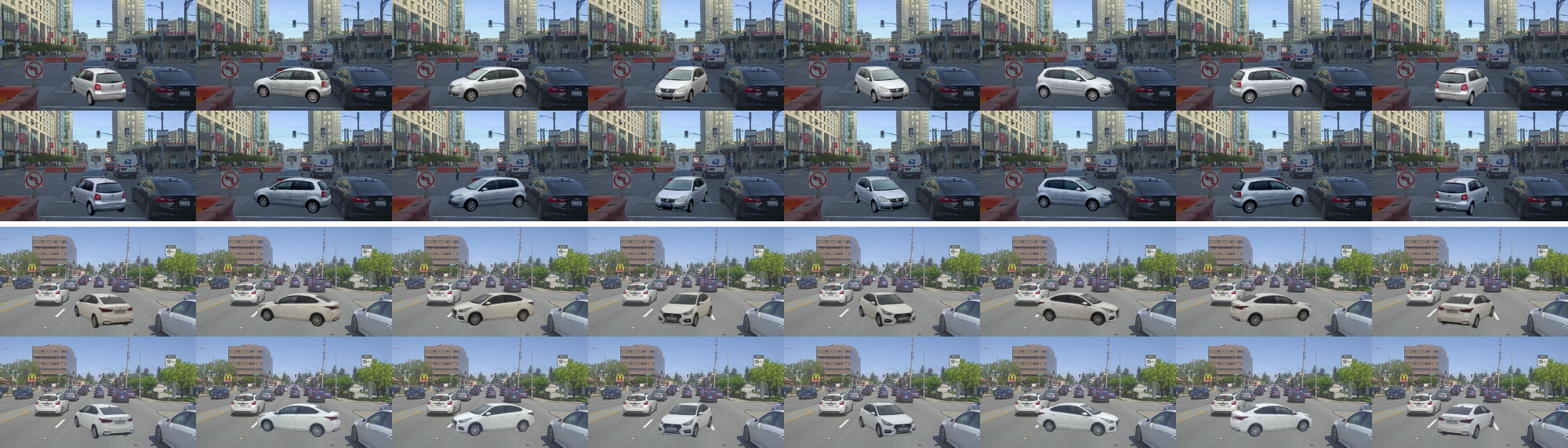}
    \caption{Qualitative comparison of harmonization results. We show multi-view renderings of the inserted assets before and after harmonization. For each scene, the first row presents the original insertion results, and the second row shows the harmonized results. By predicting assets with canonical pose and real-world scale, our model enables fully automatic insertion and harmonization without manual intervention. Across multiple views, the harmonized assets demonstrate improved consistency with surrounding environments.}
    \label{fig:harm}
    \vspace{-0.2cm}
\end{figure*}

\subsection{Gaussian Model Decoder}

Given $N$ texture tokens ${\textbf{T}}_{i=1}^N$, we employ separate MLP-based decoding heads for different Gaussian attributes to generate the corresponding properties. Each token ${\textbf{T}}_i$ decodes $m$ Gaussians primitives within a fixed spatial neighborhood around its initial position $p_i$. Specifically, for each Gaussian, the network predicts an offset relative to $p_i$ to obtain the mean position $\mu_j$. To stabilize training, the scaling $s$ and opacity $\alpha$ are also predicted as offsets from predefined bias values. Formally, the decoding process for token $\textbf{T}_i$ can be expressed as:
\begin{equation}
\lbrace \Delta p_j, \Delta s_j, \Delta \alpha_j, c_j, q_j \rbrace_j^m = \text{MLP}(\mathbf{T}_i)_j.
\end{equation}
Here, $j$ indexes the $m$ Gaussians decoded from each token, and $\text{MLP}$ denotes the separate decoding heads that predict offsets for position, scaling, opacity, as well as the spherical harmonic color $c$ and rotation $q$.


The same instance often exhibits markedly different texture appearances under varying illumination conditions. To improve the generalization and reusability of the generated assets, we introduce a style harmonization module that adapts generated model to diverse environments. We fine-tune a one-step 2D diffusion model~\cite{parmar2024one,dipir,r3d2} as a harmonization supervisor, which is trained to transform foreground-inconsistent images into visually coherent and harmonized counterparts.

Specifically, we freeze all attributes except the spherical harmonic color $c$, which primarily controls appearance, and introduce a self-attention modulation block to compute its adjustment. Given the geometry-aware texture feature $\textbf{T}_i$ of each Gaussian group, the modulation is formulated as $\lbrace\Delta c_j\rbrace_j^m = \text{Harm}(\textbf{T}_i)$, where $\Delta c_j$ denotes the adjustment for appearance. And the final SH color $c_j' = c_j + \Delta c_j$.

Benefiting from the metric-scale reconstruction and normalized spatial alignment, our generated assets can be seamlessly integrated into real-world scenes for rendering. Specifically, we construct driving scenes from PandaSet~\cite{xiao2021pandaset}, preserving the camera poses at selected timestamps, together with the ego trajectory and frames free from foreground occlusions. The generated assets are rendered using the ego poses provided in the dataset and composited with the corresponding background images, producing spatially consistent renderings without any manual adjustment. The generated assets are further refined using the appearance-harmonized results generated by the pretrained diffusion model.

\subsection{Training}

\begin{table*}[t]
\centering
\setlength{\tabcolsep}{4pt} 
\begin{tabular}{lccccc|ccccc}
\toprule
& \multicolumn{5}{c|}{\textbf{3DRealCar}} & \multicolumn{5}{c}{\textbf{CFV}} \\
\cmidrule(lr){2-6} \cmidrule(lr){7-11}
View Num & SSIM↑ & PSNR↑ & LPIPS↓ & CD↓ & F-score↑ & SSIM↑ & PSNR↑ & LPIPS↓ & CD↓ & F-score↑ \\
\midrule
1     & {0.9172} & {21.2033} & {0.0571} &{0.5782} &{0.5341} & {0.9183} & {21.7250} & {0.0508} &{0.7439} &{0.4252} \\
2     & {0.9182} & {21.4407} & {0.0548} &{0.5672} &{0.5389} & {0.9196} & {21.9673} & {0.0483} &{0.7410} &{0.4264} \\
4     & {0.9186} & {21.5565} & {0.0537} &{0.5648} &{0.5400} & {0.9203} & {22.1001} & {0.0469} &{0.7382} &{0.4291} \\
6     & {0.9187} & {21.5976} & {0.0533} &{0.5637} &{0.5406} & {0.9206} & {22.1426} & {0.0465} &{0.7351} &{0.4314} \\
8     & \textbf{0.9188} & {21.6194} & {0.0531} &\textbf{0.5630} &{0.5404} & \textbf{0.9207} & {22.1682} & {0.0462} &{0.7329} &{0.4333} \\
10     & \textbf{0.9188} & \textbf{21.6249} & \textbf{0.0529} &{0.5636} &\textbf{0.5410} & \textbf{0.9207} & \textbf{22.1737} & \textbf{0.0461} &\textbf{0.7313} &\textbf{0.4346} \\
\bottomrule
\end{tabular}
\caption{Multi-view comparison on 3DRealCar and CFV datasets. To validate the effectiveness of multi-view input, we report metrics using different numbers of input viewpoints. We randomly select $n$ views as input and evaluate the resulting models across all viewpoints.}
\label{tab:multi}
\vspace{-0.2cm}
\end{table*}

Our model is trained under pure image supervision. To enable the model to handle both multi-view and single-view inputs effectively, we adopt a probabilistic multi-view sampling strategy during training. Specifically, given $V$ available views for each scene, we initially employ all views as input and gradually reduce the probability of using multiple views in later training stages through an exponential decay schedule.

Let input view number be denoted as $v \in \lbrace{1, 2, \dots, V}\rbrace$. The unnormalized probability for selecting $v$ views is defined as $\pi_v = \exp(-\lambda (v-1))$, where $\lambda$ is the decay rate controlling how quickly the probability decreases with respect to the number of views. The normalized sampling probability is then obtained by:
\begin{equation}
\tilde{\pi}_v = \frac{\pi_v}{\sum_{i=1}^{V} \pi_i}.
\end{equation}

During each training iteration, the number of input views $v$ is sampled according to this distribution:
\begin{equation}
v \sim \text{Categorical}(\tilde{\pi}_1, \tilde{\pi}_2, \dots, \tilde{\pi}_{V}).
\end{equation}
This strategy allows the model to maintain its single-view reconstruction capability while enhancing its ability to aggregate information from multiple views.


As described in the previous sections, our method follows a two-stage training paradigm. In the first stage, the model is pre-trained on the camera-annotated dataset 3DRealCar~\cite{3drealcar}, and subsequently, it is generalized to a larger-scale unposed image dataset MAD-Cars~\cite{madrive}.

To prevent the degeneration of Gaussian primitives and ensure that each Gaussian contributes meaningfully to the final model, we introduce regularization terms on opacity and scaling. Specifically, given the predicted opacity $\alpha_i$ and scaling factor $s_i$ of the $i$-th Gaussian, the regularization terms are formulated as:
\begin{equation}
\mathcal{L}_{\text{reg}} = \frac{1}{N} \sum_{i=1}^{N} (\alpha_i - 1)^2 + \frac{1}{N} \sum_{i=1}^{N} \max_j s_{i,j}, 
\end{equation}
where $N$ denotes the total number of Gaussians, and $s_{i,j}$ is the scaling along the $j$-th axis of the $i$-th Gaussian.

The overall loss in the first stage combines image reconstruction and perceptual terms, camera supervision, and the aforementioned regularization:
\begin{equation}
\mathcal{L}_{1}=\mathcal{L}_{\text{L1}}+\mathcal{L}_{\text{SSIM}}+\mathcal{L}_{\text{LPIPS}}+\mathcal{L}_{\text{scale}}+\mathcal{L}_{\text{cam}}+\mathcal{L}_{\text{reg}}.
\end{equation}


In the second training stage, we rely solely on image-level reconstruction losses, which jointly optimize both the generative quality and the accuracy of camera parameter prediction. This introduces several challenges. First, since both camera parameters and 3D geometry are learnable, a distant camera with a narrow FOV or distorted geometry with inaccurate intrinsics can produce visually correct results. We refer to this as the geometry-camera ambiguity problem. Second, small errors in camera prediction can lead to big pixel-level misalignments, resulting in the loss of high-frequency details. Finally, when pretraining and fine-tuning on datasets with different image and camera distributions, domain gaps make training convergence difficult.

To mitigate this, we propose a gradient filtering strategy across second training stages. Specifically, we treat a batch of data as a sliding window and compute the gradient contributions of each view. Gradients that deviate beyond $3\sigma$ are discarded. Gradients that are too small indicate the model has already sufficiently fitted that view, and further optimization risks overfitting. Gradients that are too large indicate severe camera misprediction, and backpropagating them could harm optimization. Over the course of training, as camera predictions become more accurate, previously discarded outlier training data are reincorporated. This strategy ensures supervision signals are applied only when rendered views align well with the target, while preventing harmful gradients caused by geometry-camera ambiguity, allowing stable cross-dataset training.

Formally, given per-view gradient contributions $L_i$ with mean $\bar{L}$ and standard deviation $\sigma_L$, we define the gradient filtering mask as $M_i = \mathbf{1}[\bar{L}-3\sigma_L < L_i < \bar{L}+3\sigma_L]$. The masked gradients used for backpropagation are then computed as:
\begin{equation}
\tilde{\nabla}_\theta = \sum_{i=1}^{V} M_i \nabla_\theta L_i,
\end{equation}
where $\tilde{\nabla}_\theta$ denotes the filtered gradient, ensuring that only reliable contributions influence the model update. 

In the second stage, we remove the supervision of camera parameters and scale, retaining only image-level supervision and regularization constraints:
\begin{equation}
\mathcal{L}_{2}=\mathcal{L}_{\text{L1}}+\mathcal{L}_{\text{SSIM}}+\mathcal{L}_{\text{LPIPS}}+\mathcal{L}_{\text{reg}}.
\end{equation}

After completing the second training stage, we decouple the scale prediction head and finetune it for a few epochs on the 3DRealCar~\cite{3drealcar} dataset to regress the real-world scale.

\section{Experiments}

\subsection{Datasets}
The 3DRealCar dataset~\cite{3drealcar} provides densely sampled multi-view images surrounding each vehicle, accompanied by accurate camera annotations and metric-scale calibration, enabling reconstruction with real-world scale. It contains approximately 2,500 samples in total. We select about 1,000 high-quality samples for the first training stage and reserve 100 samples for evaluation. The MAD-Cars dataset~\cite{madrive} contains around 70,000 vehicle instances and approximately 5 million images without camera annotations. We use MAD-Cars in the second training stage to enhance the generalization capability of our model. The CFV dataset~\cite{cfv} includes 65 vehicle instances, which we employ as the test set for zero-shot generalization experiments.

\subsection{Single View Reconstruction}
To validate the effectiveness of our method for single-view reconstruction, we conducted both quantitative and qualitative comparisons on the 3DRealCar and CFV datasets. As shown in \autoref{fig:show}, the qualitative results reveal that DGS~\cite{dreamgaussian} tends to produce blurry textures and loses substantial high-frequency details. TGS~\cite{tgs} and LGM~\cite{lgm} often overfit to the input views, failing to generalize to novel viewpoints and being prone to reconstruction collapse. TRELLIS~\cite{trellis} frequently generates models with abnormal deformations, and its outputs exhibit a strong simulation-style appearance. In contrast, our method consistently preserves complete and realistic geometry while producing rich, detailed textures across diverse scenarios, with texture fidelity substantially higher than that of the baseline methods. 

We also provide qualitative results for our harmonization method in \autoref{fig:harm}. Directly inserting reconstructed assets into novel scenes often leads to inconsistencies in lighting and color, which significantly reduces the realism of the simulation. After applying harmonization modulation, the generated assets better adapt to the corresponding scenes, enhancing visual realism and facilitating the generalized use of the assets across different scenarios.

The quantitative results are presented in \autoref{tab:single}. The evaluation procedure involves reconstructing a model from a single image and computing the metrics across all viewpoints. For a fair comparison, all generated models were first normalized to a standard scale, and the poses of the reconstructed assets from all baseline methods were manually adjusted to eliminate metric errors caused by differing model poses. This also highlights the limited usability of the baseline methods, whereas our approach requires no additional manual adjustment. Here, Chamfer Distance (CD) is scaled by 1000, and the F-score is computed with a radius of 0.01 in the normalized scale. As shown in \autoref{tab:single}, our method outperforms all baseline methods in both 2D image reconstruction metrics and 3D reconstruction metrics.

\begin{table}[t]
    \centering
    \begin{tabular}{lccc}
        \toprule
        Dataset & SSIM $\uparrow$ & PSNR $\uparrow$ & LPIPS $\downarrow$ \\
        \midrule
        3DRealCar   & 0.9263 & 21.7589 & 0.0500 \\
        CFV         & 0.9239 & 21.7762 & 0.0477 \\
        \bottomrule
    \end{tabular}
    \caption{Quantitative evaluation of camera pose estimation accuracy. We use single image as input and render with the predicted camera parameters, then compute metrics against the ground-truth images to evaluate the accuracy of camera parameter estimation.}
    \label{tab:pose-metrics}
    \vspace{-0.2cm}
\end{table}

\begin{table}[t]
    \centering
    \begin{tabular}{lcc}
        \toprule
        Dataset & CD (m) $\downarrow$ & F-Score $\uparrow$ \\
        \midrule
        3DRealCar & 0.0160 & 0.4886 \\
        \bottomrule
    \end{tabular}
    \caption{Evaluation of metric-scale reconstruction accuracy on real-world datasets. The Chamfer Distance (CD) is measured in meters, and F-score is computed with a threshold of $0.05$ m.}
    \label{table:scale}
    \vspace{-0.2cm}
\end{table}

\subsection{Multi View Reconstruction}
To validate the effectiveness of our multi-view input design, we conduct quantitative experiments using varying numbers of input views. Specifically, we randomly select $n$ views as input and evaluate the metrics across all available viewpoints. The results are presented in \autoref{tab:multi}. It can be observed that as the number of input views increases, both visual and geometric quality improve. However, the rate of improvement diminishes with more views, as additional inputs eventually provide sufficient prior information to the model. The differences in metrics across different numbers of input views are relatively small, which can be attributed to the random selection of views. For example, if the chosen views are mostly from the same side, they may offer limited additional information to the model.

\subsection{Camera Prediction and Scale}
The capability of camera prediction and the accuracy of real-world scale are unique to our method, therefore, only our model is evaluated. To assess the accuracy of our camera prediction module, we leverage image reconstruction loss as an indicator. Specifically, for each ground-truth image, our method simultaneously predicts the 3D model and camera parameters. The predicted camera parameters are then used to render the model, and the reconstruction loss is computed against the input image. This design ensures that image reconstruction metrics reflect the reliability of both the reconstructed model and the predicted camera parameters. As shown in \autoref{tab:pose-metrics}, the reconstruction metrics computed using our predicted camera parameters achieve strong quantitative results across both datasets, demonstrating the high precision of our camera prediction module.

To evaluate the accuracy of our generative model in recovering real-world scale, we report metrics computed at the true metric scale, as shown in \autoref{table:scale}. Since the CFV dataset does not provide scale information, only results on 3DRealCar are presented. Here, Chamfer Distance is measured in meters, and F-score is evaluated at a radius of $0.05$ m. These results demonstrate that our model achieves high geometric reconstruction accuracy at real-world scale.

\begin{table}[h]
\centering
\begin{tabular}{@{}lcccc@{}}
\toprule
Setting & L1/AP$\uparrow$ & L1/APH$\uparrow$ & L2/AP$\uparrow$ & L2/APH$\uparrow$ \\
\midrule
Real & 0.8837 & 0.8594 & 0.8026 & 0.7793 \\
Real + Sim & \textbf{0.8946} & \textbf{0.8738} & \textbf{0.8198} & \textbf{0.7996} \\
\bottomrule
\end{tabular}
\caption{To evaluate the utility of the generated 3D assets, we perform scene augmentation to boost the performance of downstream 3D object detection.}
\label{tab:sim2real}
\vspace{-0.2cm}
\end{table}

\subsection{Sim2real Downstream Validation}
To substantiate the simulation-ready claim, we evaluate our assets in the 3D object detection task using the OpenPCDet~\cite{openpcdet2020} framework with Voxel R-CNN~\cite{deng2021voxel} as the detector. Specifically, we perform data augmentation by inserting our generated assets into 100 randomly selected Waymo~\cite{waymo} scenes. Following the standard experimental setup for 3D object detection~\cite{openpcdet2020}, we observe steady improvements in $\rm{AP}$ and $\rm{APH}$ across both $\rm{LEVEL\_1}$ and $\rm{LEVEL\_2}$ difficulties, as detailed in ~\autoref{tab:sim2real}.

\section{Conclusion}

We present Unposed-to-3D, a fully image-driven framework that reconstructs simulation-ready vehicles from real-world data. Supporting arbitrary view inputs, our method enables self-supervised camera pose estimation and metrically accurate 3D reconstruction without any 3D ground truth. Moreover, the proposed lightweight harmonization module ensures that reconstructed vehicles can be seamlessly integrated into diverse environments with varying illumination and scene contexts. Extensive experiments show that our approach yields scale-consistent, pose-accurate, and visually coherent 3D vehicles, significantly enhancing the realism and usability of real-world simulations. 

\noindent{\bf{Acknowledgement}} This work was supported by the Beijing Natural Science Foundation (No. L257003, 4262058), National Natural Science Foundation of China (No. 62227801, 62576032), Fundamental Research Funds for the Central Universities (No. FRF-KST-25-008), and Young Scientist Program of The National New Energy Vehicle Technology Innovation Center (Xiamen Branch).

{
    \small
    \bibliographystyle{ieeenat_fullname}
    \bibliography{main}
}

\clearpage
\setcounter{page}{1}
\maketitlesupplementary

\newcommand{\minisection}[1]{%
  \vspace{0.75em}%
  \noindent\textbf{#1}~
}

\section{Implementation Details}

\subsection{Network Architectures}

\begin{figure}[h]
    \centering
    \includegraphics[width=\linewidth]{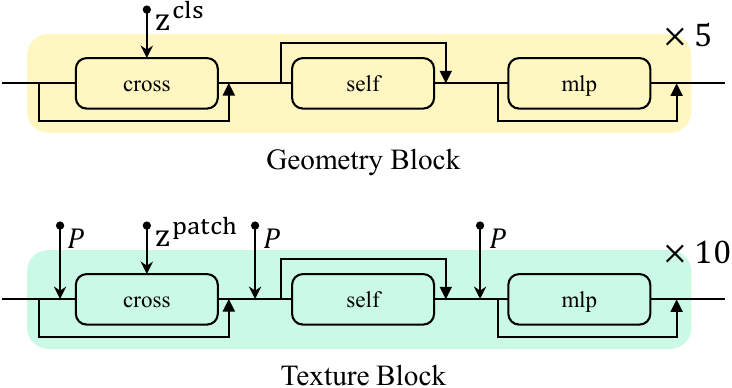}
    \caption{The two base modules of the backbone network.}
    \label{fig:block}
\end{figure}



\minisection{Aggregation Module.}We adopt the DINOv2-base~\cite{dinov2} encoder and uniformly resize all input images to $224 \times 224$. A learnable camera embedding is introduced as a single token without positional encoding. The aggregation module consists of four Alternating Self-Attention blocks, each containing two self-attention layers. An attention-pooling layer implemented with a single-layer MLP predicts view-dependent weighting coefficients over all tokens and produces the final aggregated feature representation.

\minisection{Learnable Object Embedding.}We parameterize the structural representation as a learnable embedding of shape $N \times C$, where $N = 4096$ and $C = 768$, matching the dimensionality of DINO features. Each token is equipped with a sinusoidal positional encoding defined as:
\begin{equation}
\text{PE}(x) = [\sin(x \cdot f_i), \cos(x \cdot f_i)]_{i=1}^{C/2},\quad
f_i = \frac{1}{10000^{i/(C/2)}},
\end{equation}
where $x \in \{{0,1,\dots,N-1\}}$ denotes the token index. This encoding provides spatial inductive bias for the learnable object tokens.

\minisection{Backbone.}As illustrated in \autoref{fig:block}. The Geometry Block is composed of five blocks. It outputs geometry features $G$, which are decoded by three single-layer fully connected heads to predict spherical coordinates $(r, \theta, \phi)$. These are then converted into Cartesian coordinates $P=(x, y, z)$. The predicted 3D positions are subsequently used inside the Texture Blocks to modulate per-token features, injecting explicit spatial information into the texture representation. The Texture Block contains ten base blocks. The resulting feature tokens are decoded by zero-initialized single-layer linear heads into $m = 8$ Gaussian attributes per token. Given $N = 4096$ tokens, this results in a total of $8 \times 4096 = 32768$ 3D Gaussians. 

\minisection{Training.}We train the entire network using AdamW with an initial learning rate $\text{lr=0.00016}$. A warm-up phase linearly ramps the learning rate from $0.1 \times \text{lr}$ over the first $T_{\text{warmup}}$ iterations. This is followed by a cosine annealing schedule that decays the learning rate to $0.01 \times \text{lr}$ over the remaining training steps. Our experiments are conducted on 24 NVIDIA A800 GPUs. The full training process takes approximately three days. We first train the model for 30 epochs on the 3DRealCar~\cite{3drealcar} dataset, followed by an additional 5 epochs of fine-tuning on the MAD-Cars~\cite{madrive} dataset.

\subsection{Data Preparation Details}

\begin{figure}[h]
    \centering
    \includegraphics[width=0.7\linewidth]{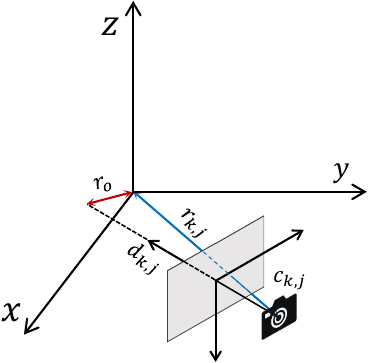}
    \caption{Schematic illustration of the camera construction}
    \label{fig:random}
\end{figure}

\begin{figure*}[t]
    \centering
    \captionsetup{type=figure}
    \includegraphics[width=\textwidth]{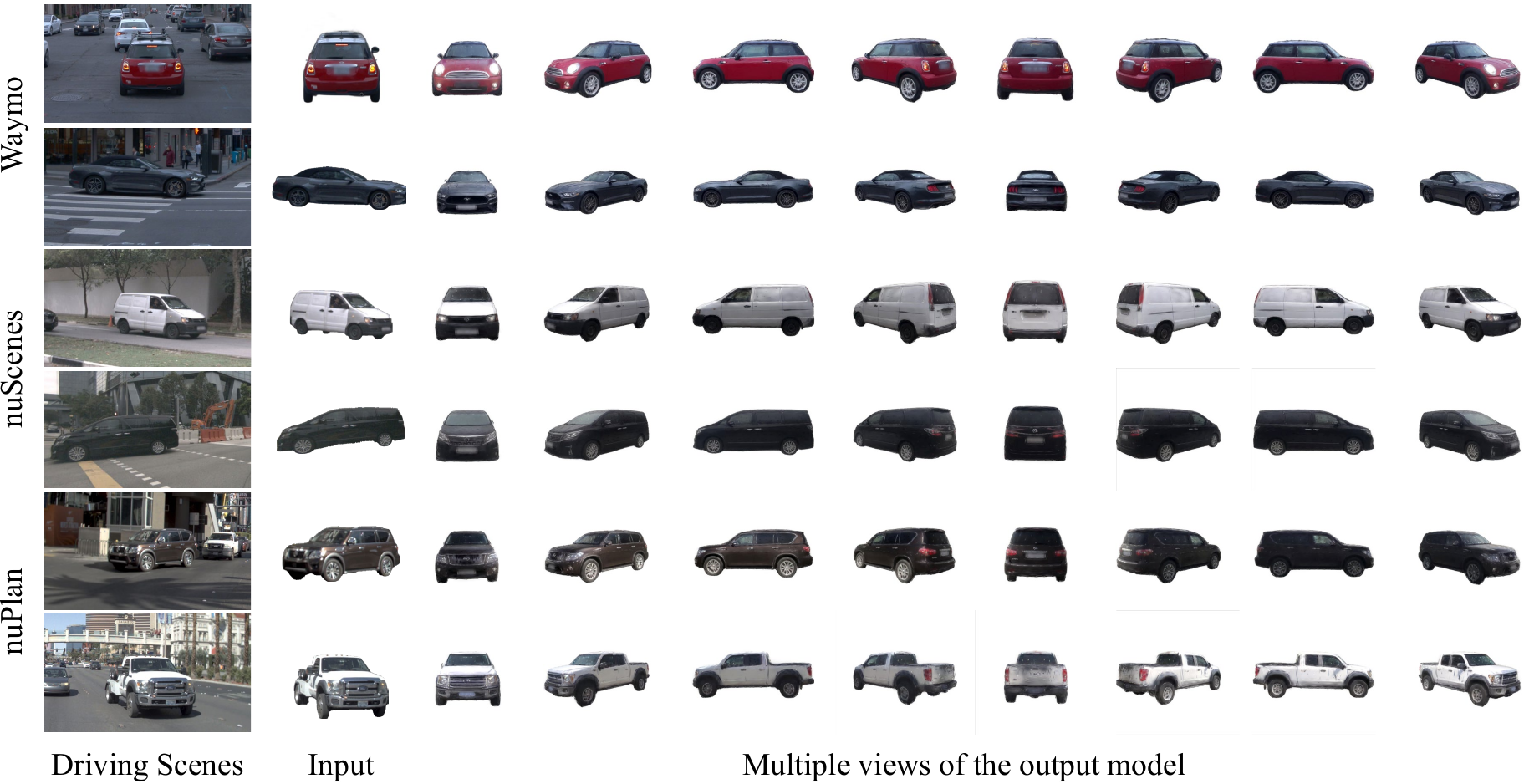}
    \caption{Qualitative generalization results of single-image reconstruction on autonomous driving datasets.}
    \label{fig:drive_data}
\end{figure*}

\minisection{3DRealCar.}To generate training views for each object instance, we place virtual cameras on randomized spherical trajectories surrounding the normalized 3D model. We sample $T=4$ orbits, each containing $V=32$ equally spaced azimuth angles. For the $k$-th orbit, the elevation angle $\phi_k$ is drawn from a uniform range $[75^\circ, 90^\circ]$, and a per-view camera radius $r_{k,j}\sim\mathcal{U}(1.2, 1.8)$ is used to introduce additional geometric diversity. The spherical position of the $j$-th camera on orbit k is given by
\begin{equation}
\mathbf{c}_{k,j}=
\begin{bmatrix}
r_{k,j}\sin\phi_k\cos\theta_{k,j} \\
r_{k,j}\sin\phi_k\sin\theta_{k,j} \\
r_{k,j}\cos\phi_k
\end{bmatrix},
\theta_{k,j}=\frac{2\pi j}{V} + \delta_j ,
\end{equation}
where $\delta_j$ is a small random offset to avoid perfectly symmetric camera layouts. Rather than enforcing a fixed look-at direction, each camera is oriented toward a randomly sampled point:

\begin{equation}
\hat{{v}}_i = \frac{{v}_i}{|{v}_i|_2}, {v}_i \sim \mathcal{N}(\mathbf{0}, \mathbf{I}_3),
\end{equation}
where $i=\{1,2,\dots,TV\}$. The points $p_{k,j}$ are sampled within a small sphere centered at the origin, with the radius $r_o \sim \mathcal{U}(0,0.1)$, i.e., $p_{k,j} = r_o \hat{v}_{ij}$. This defines the viewing direction as

\begin{equation}
d_{k,j}=\frac{{p}_{k,j}-{c}_{k,j}}{|{p}_{k,j}-{c}_{k,j}|},
\end{equation}
which induces natural variations in camera orientation, thereby mitigating overfitting. The camera intrinsics are also randomized for each orbit: a field of view $\mathrm{FOV}\sim\mathcal{U}(50^\circ,70^\circ)$ is sampled, while the principal point remains fixed at the image center. For each pair of extrinsic parameters $(\mathbf{q},\mathbf{t})$ and intrinsics $\mathbf{f}$, we render a $512\times 512$ RGB image along with its corresponding alpha mask.

\minisection{MAD-Cars.}The MAD-Cars~\cite{madrive} dataset contains only image data. For each image, we first extract instance-level bounding boxes and masks using YOLO~\cite{Jocher_Ultralytics_YOLO_2023} and MobileSAM~\cite{mobile_sam}. We then discard images whose bounding boxes lie near the image boundaries, as such crops are likely to be incomplete. For each remaining instance, we normalize all associated images by scaling them according to the longest side of its bounding box and centering the bounding box within the image.

\section{More Results}

\minisection{Zero-shot in Driving Scenario.}
As shown in \autoref{fig:drive_data}, we provide additional qualitative results on autonomous driving datasets. Relying solely on single-view inputs, our method is able to recover high-quality vehicle models from real driving environments, demonstrating its strong generalization capability and practical applicability.

\minisection{More Comparisons.}To facilitate a qualitative comparison between the proposed method and the baselines, we present additional qualitative comparison on the 3DRealCar~\cite{3drealcar} and CFV~\cite{cfv} datasets, as shown in \autoref{fig:supp_show_realcar_1}, \autoref{fig:supp_show_realcar_2}, \autoref{fig:supp_show_CFV_1}, \autoref{fig:supp_show_CFV_2}. Across varying lighting conditions and diverse asset types, our method consistently outperforms the baseline approaches.

\minisection{Camera Estimation.}To validate the accuracy of our camera parameter predictions, we provide qualitative results in \autoref{fig:supp_cam_show}. For each instance, the first row shows the input ground truth images without camera poses, and the second row presents the rendered outputs obtained by applying the predicted camera parameters to our reconstructed models. As observed, our camera predictions achieve pixel-level precision, which is crucial for enabling high-quality 3D reconstruction using only unposed image supervision. This capability, absent in prior 3D generative models, highlights the scalability of our method.

\begin{figure*}[t]
    \centering
    \captionsetup{type=figure}
    \includegraphics[width=\textwidth]{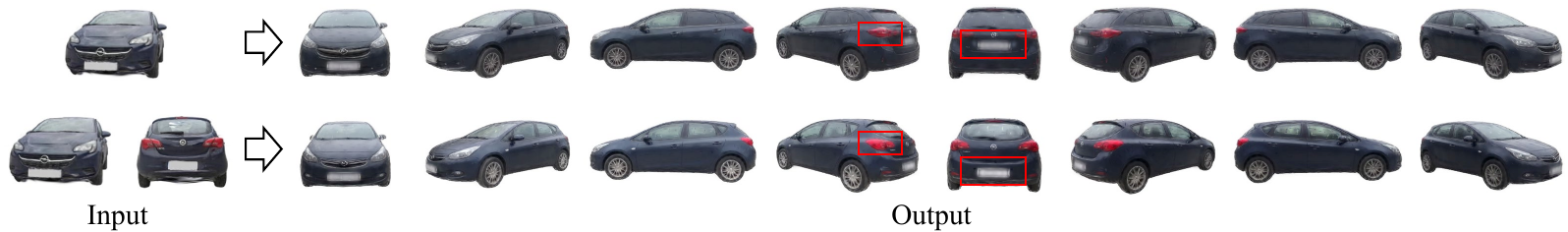}
    \caption{Qualitative reconstruction results under different input settings. We present results using a single input view as well as using two input views, where an additional viewpoint is provided to enhance reconstruction quality.}
    \label{fig:mv}
\end{figure*}

\begin{figure*}[t]
    \centering
    \captionsetup{type=figure}
    \includegraphics[width=\textwidth]{fig/supp_mesh.pdf}
    \caption{We visualize the Gaussian points to illustrate geometric quality. Compared to TRELLIS~\cite{trellis}, our method achieves high-fidelity geometric reconstruction using substantially fewer Gaussian points.}
    \label{fig:mesh}
\end{figure*}

\begin{figure}[h]
    \centering
    \includegraphics[width=\linewidth]{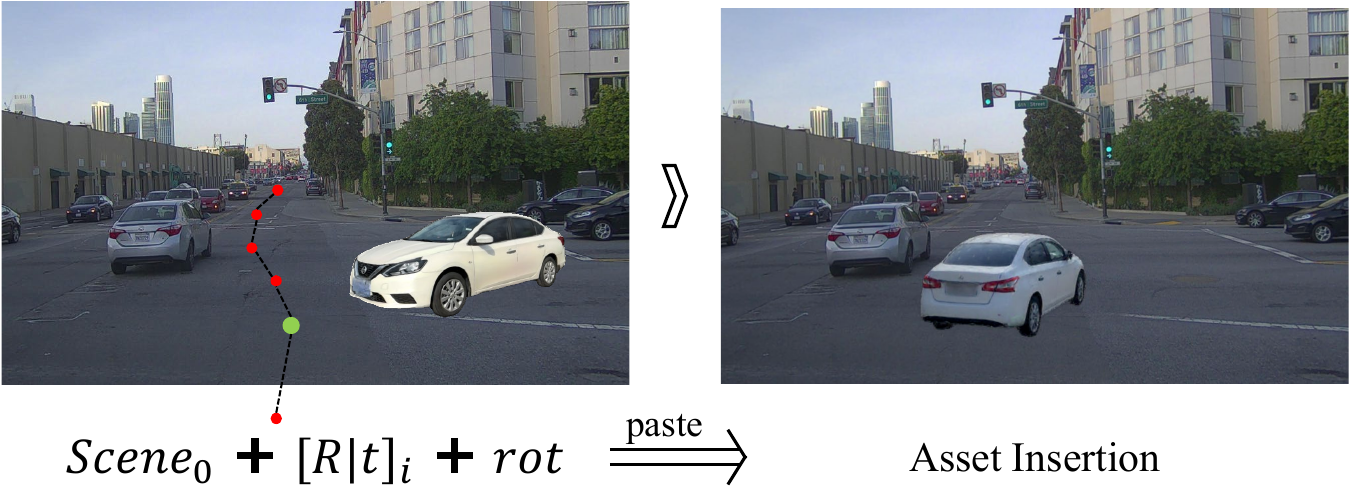}
    \caption{Harmonized Training Scene Construction.}
    \label{fig:harm_scene}
\end{figure}

\minisection{Multi View Reconstruction.}To validate the effectiveness of our multi-view input strategy, we present qualitative comparisons in \autoref{fig:mv}. The first row shows reconstruction results using only a frontal-view image, while the second row incorporates an additional rear-view image as input. Adding an extra viewpoint improves the reconstruction quality of rear details, such as taillights and license plates, demonstrating the advantages of multi-view input in our method. Because the single-view setting already yields strong reconstruction quality, incorporating multi-view inputs only further refines some fine-grained geometric details, so the resulting quantitative improvements remain marginal, which is consistent with the main paper.

\minisection{Geometric Reconstruction.}To evaluate the geometric reconstruction quality of our method, we visualize the generated Gaussian point clouds in \autoref{fig:mesh}. The distribution of Gaussians produced by our model closely aligns with the underlying image textures: high-frequency regions are represented with a higher density of Gaussian points, whereas low-frequency areas require substantially fewer points. In contrast, the baseline method TRELLIS~\cite{trellis} typically relies on hundreds of thousands of points to represent a single object. Remarkably, our approach achieves superior reconstruction quality using only about thirty thousand Gaussians, demonstrating a compact yet more expressive geometric representation.

\minisection{Harmonization.}The module consists of a single self-attention block followed by a two-layer MLP, and is trained under $\text{MSE}$, $\text{SSIM}$, $\text{LPIPS}$, and $\text{L}_1$ losses to jointly regularize appearance. The learning rate is set to 0.1, and the weights of all image-level losses are fixed to 10. As shown in \autoref{fig:harm_scene}, unlike prior approaches that require reconstructing the entire background and rely on manual adjustments for 3D asset insertion, we only retain the background image of the initial frame $scene_0$, along with the ego relative rotation and translation $[R|t]_t$ from the target keyframe to the initial frame. For the harmonized generated assets, we further apply an additional rotation $rot$ around the upright axis to adjust their pose and modulate the texture properties at different views. Finally, we render only the foreground object and paste it onto the background, producing a geometrically consistent scene. We provide additional visual results of harmonization across diverse scenes and assets in \autoref{fig:supp_harm}.

\begin{table*}[t]
\centering
\begin{tabular}{@{}cc|ccccc|ccccc@{}}
\toprule
\multicolumn{2}{c|}{} & \multicolumn{5}{c|}{3DRealcar} & \multicolumn{5}{c}{CFV} \\
Sphere & Filter & SSIM$\uparrow$ & PSNR$\uparrow$ & LPIPS$\downarrow$ & CD$\downarrow$ & F-score$\uparrow$ & SSIM$\uparrow$ & PSNR$\uparrow$ & LPIPS$\downarrow$ & CD$\downarrow$ & F-score$\uparrow$ \\
\midrule
\xmark & \xmark & \multicolumn{5}{c}{--} & \multicolumn{5}{c}{--} \\
\cmark & \xmark & 0.9059 & 19.8572 & 0.0742 & 0.9348 & 0.3917 & 0.9043 & 20.1337 & 0.0703 & 1.1861 & 0.2940 \\
\midrule
\cmark & $1\sigma$ & 0.9136 & 20.7738 & 0.0617 & 0.7195 & 0.4776 & 0.9125 & 21.1185 & 0.0570 & 0.9032 & 0.3455 \\
\cmark & $2\sigma$ & 0.9121 & 20.5656 & 0.0637 & 0.7339 & 0.4654 & 0.9118 & 21.0839 & 0.0571 & 0.9482 & 0.3374 \\
\cmark & $3\sigma$ & \textbf{0.9172} & \textbf{21.2033} & \textbf{0.0571} & \textbf{0.5782} & \textbf{0.5341} & \textbf{0.9183} & \textbf{21.7250} & \textbf{0.0508} & \textbf{0.7439} & \textbf{0.4252} \\
\bottomrule
\end{tabular}
\caption{Ablation study of sphere initialization and gradient filtering strategies.}
\label{tab:ablation}
\end{table*}

\minisection{More Ablation.} As shown in~\autoref{tab:ablation}, we conduct further ablation studies to evaluate our method. Training explicit geometry generation without 3D geometric supervision is a severely ill-posed problem. Without our Progressive Spherical Optimization, the network suffers from collapse and fails to produce meaningful results. $3\sigma$ Gradient Filtering strategy is grounded in the Standard Deviation Rule, where gradients exceeding $3\sigma$ are treated as outliers. This effectively prevents erroneous gradients from initial camera-geometry ambiguities.


\section{Limitations and Future works}
Our method is unable to handle input images that exhibit geometric distortions; severe non-uniform stretching or compression typically leads to corresponding deformations in the reconstructed 3D model. In addition we do not explicitly model shadows cast by environmental illumination. The realism of asset insertion can still be improved during scene integration~\cite{hugsim}.

\begin{figure*}[ht]
    \centering
    \captionsetup{type=figure}
    \includegraphics[width=0.8\textwidth]{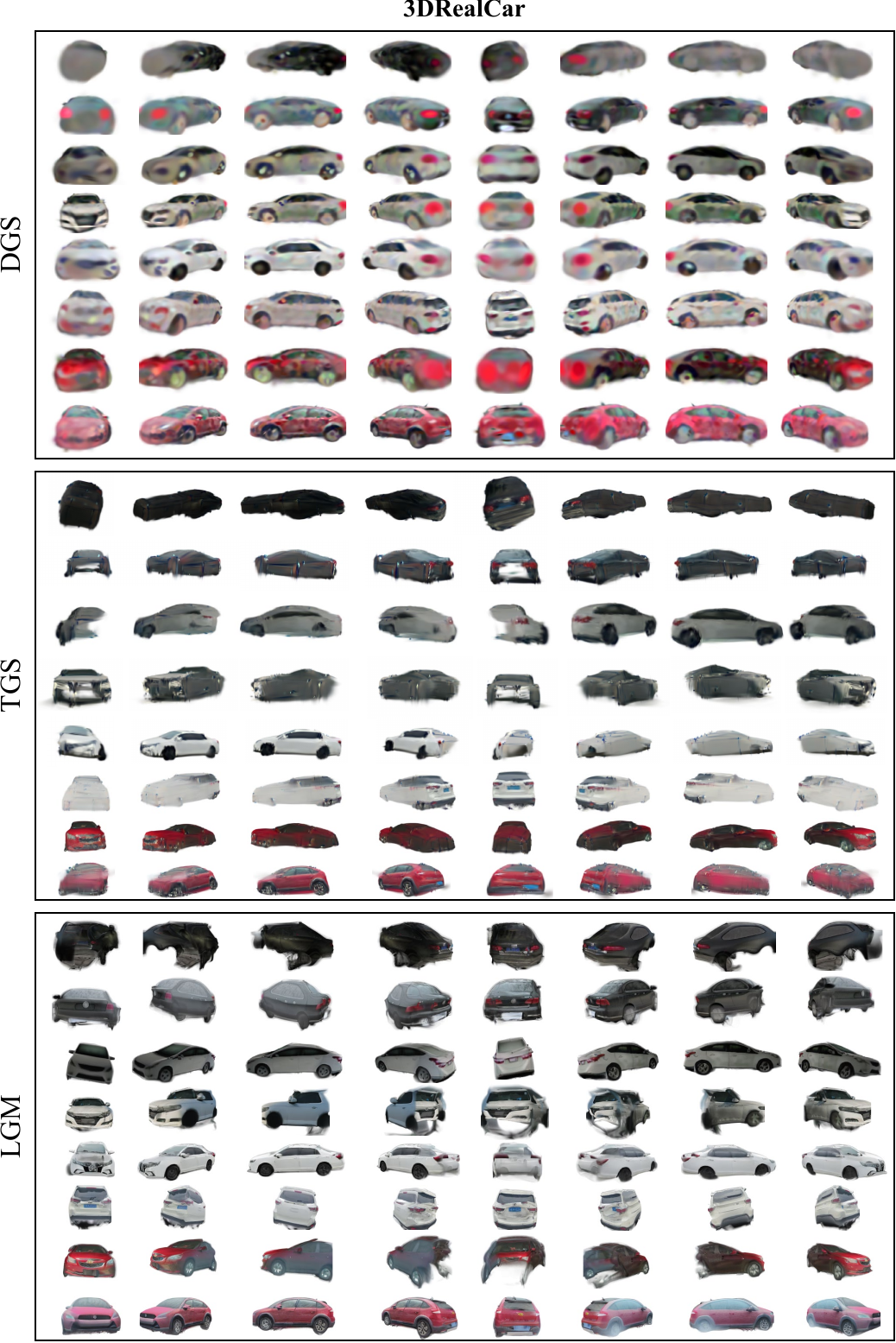}
    \caption{Qualitative results of LGM, TGS, DGS on 3DRealCar~\cite{3drealcar}.}
    \label{fig:supp_show_realcar_1}
\end{figure*}

\begin{figure*}[ht]
    \centering
    \captionsetup{type=figure}
    \includegraphics[width=0.8\textwidth]{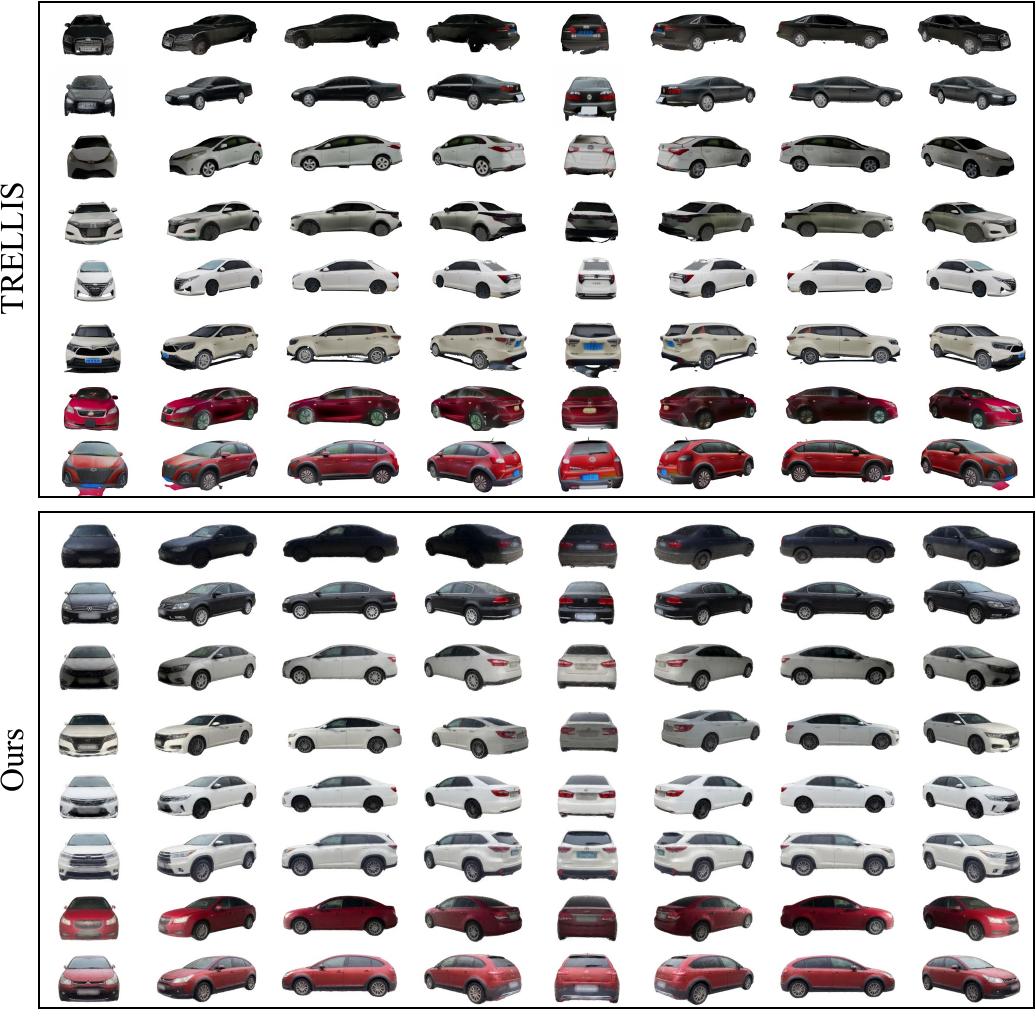}
    \caption{Qualitative results of TRELLIS and ours on 3DRealCar~\cite{3drealcar}.}
    \label{fig:supp_show_realcar_2}
\end{figure*}

\begin{figure*}[ht]
    \centering
    \captionsetup{type=figure}
    \includegraphics[width=0.8\textwidth]{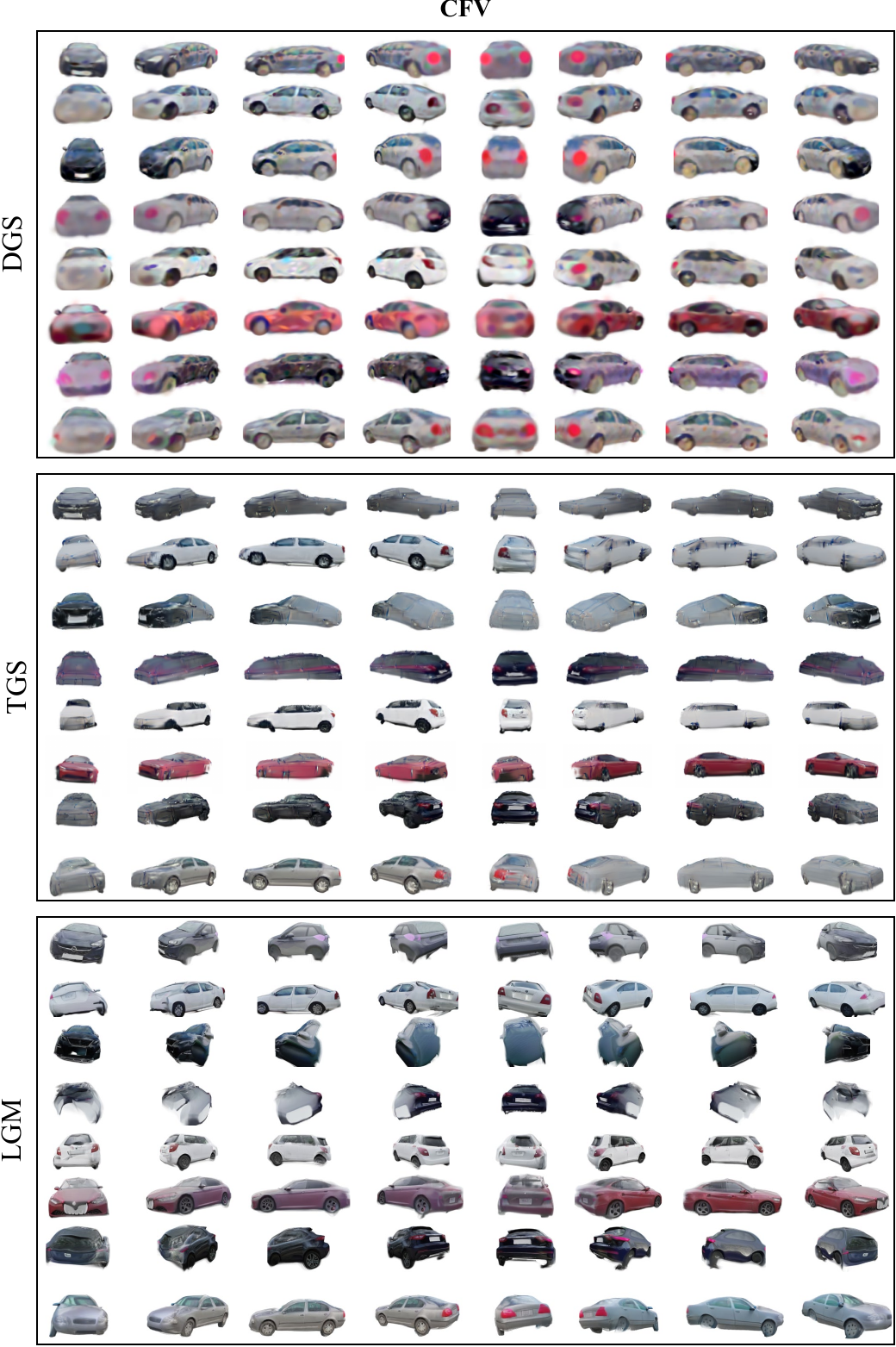}
    \caption{Qualitative results of LGM, TGS, DGS on CFV~\cite{cfv}.}
    \label{fig:supp_show_CFV_1}
\end{figure*}

\begin{figure*}[ht]
    \centering
    \captionsetup{type=figure}
    \includegraphics[width=0.8\textwidth]{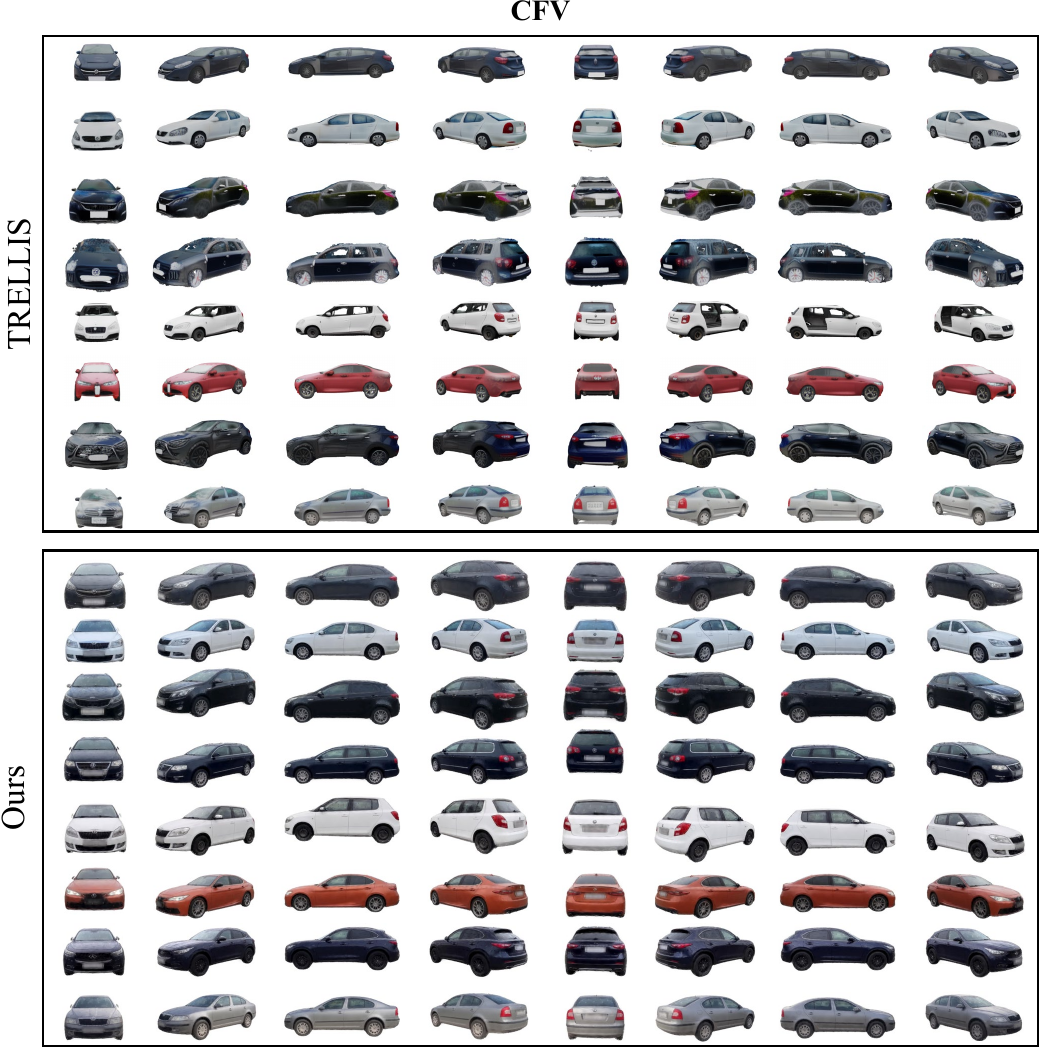}
    \caption{Qualitative results of TRELLIS and ours on CFV~\cite{cfv}.}
    \label{fig:supp_show_CFV_2}
\end{figure*}

\begin{figure*}[ht]
    \centering
    \captionsetup{type=figure}
    \includegraphics[width=\textwidth]{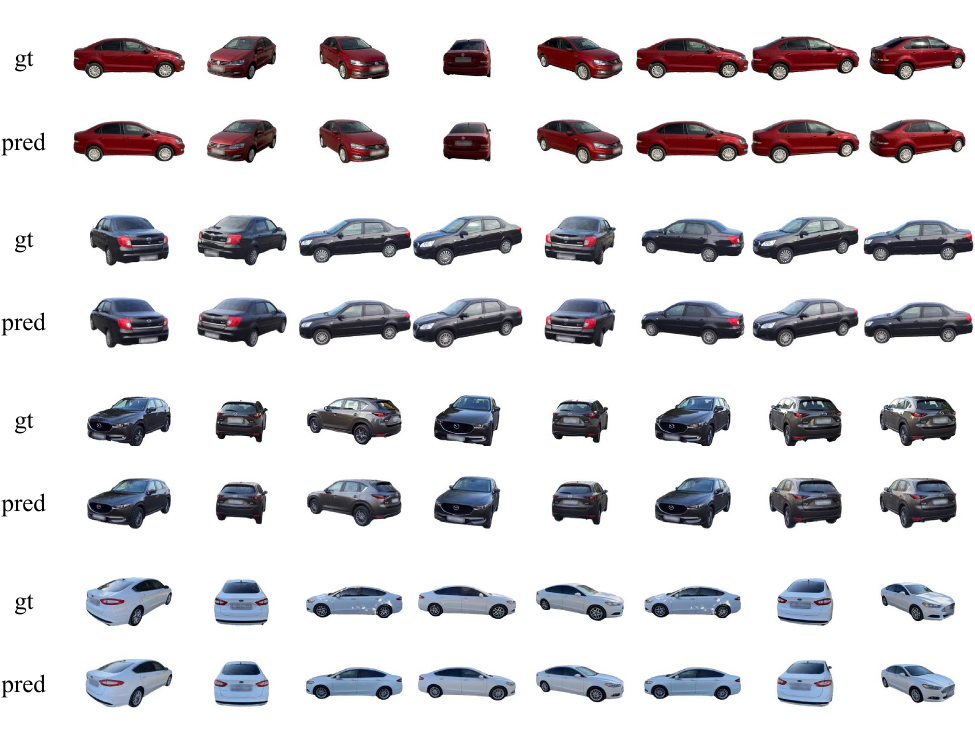}
    \caption{Qualitative results of camera parameters prediction.}
    \label{fig:supp_cam_show}
\end{figure*}

\begin{figure*}[ht]
    \centering
    \captionsetup{type=figure}
    \includegraphics[width=0.88\textwidth]{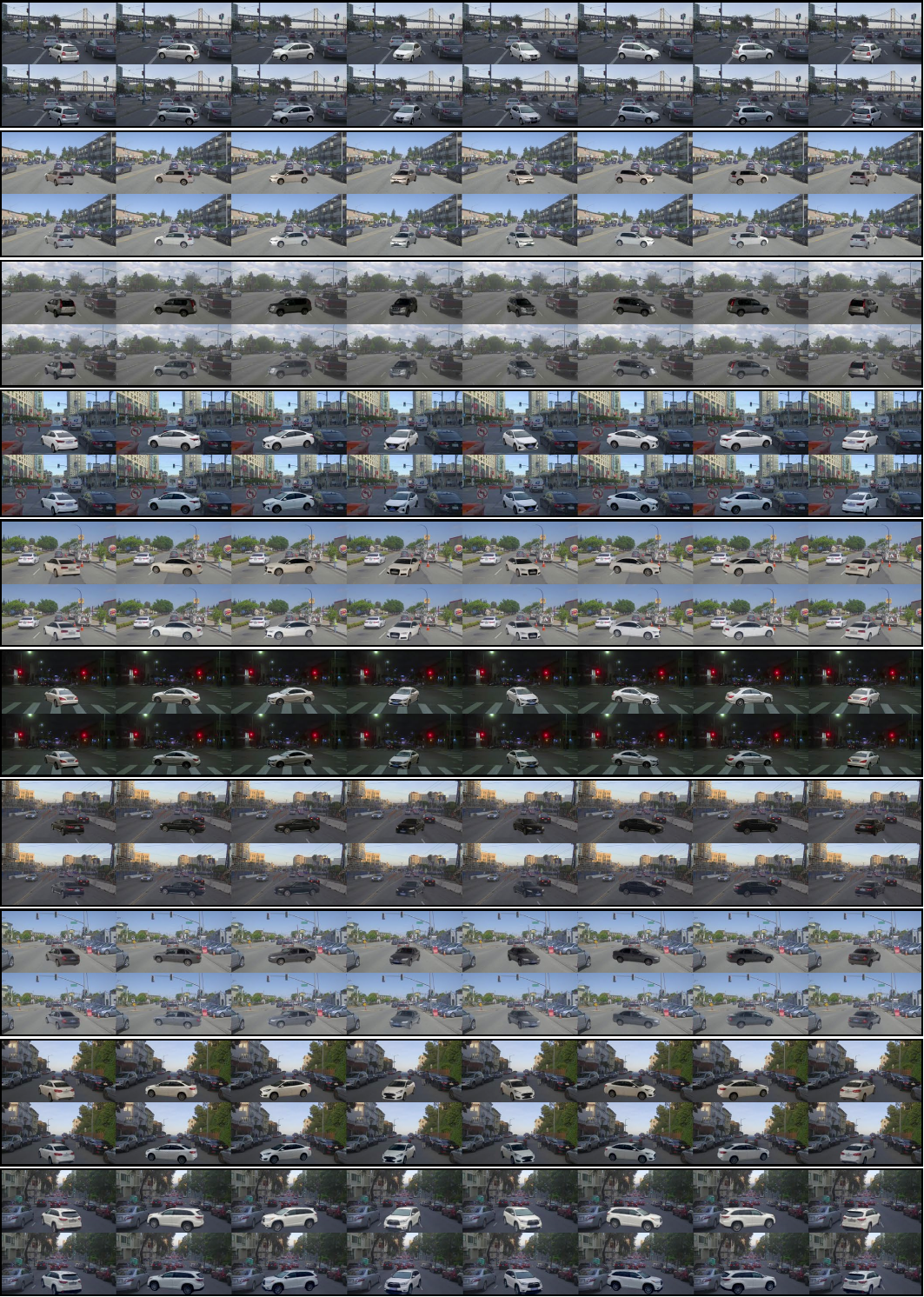}
    \caption{Qualitative results of harmonization. The first row presents the pre-harmonization outputs, while the second row shows the harmonized results. Examples are shown across multiple poses of the same asset within the scene to illustrate consistency and robustness.}
    \label{fig:supp_harm}
\end{figure*}

\end{document}